*The following manuscript is on analysis of adverse events in robotic surgical systems during the 14 year period of 2000–2013. This is an update to our analysis which was originally presented at the 50$^{th}$ Annual Meeting of the Society of Thoracic Surgeons in January 2013. Please see Appendix for more detailed results, discussions, and related work.*

# Adverse Events in Robotic Surgery:

# A Retrospective Study of 14 Years of FDA Data

Homa Alemzadeh[1], Ravishankar K. Iyer[1], Zbigniew Kalbarczyk[1], Nancy Leveson[2], Jai Raman[3]

[1]*University of Illinois at Urbana-Champaign - {alemzad1, rkiyer, kalbarcz}@illinois.edu*

[2]*Massachusetts Institute of Technology - leveson@mit.edu*

[3]*Rush University Medical Center - jai_raman@rush.edu*

**Corresponding Author:**

Jai Raman, MD FRACS PhD

1725 W Harrison St, Suite 1156

Rush University Medical Center, Chicago, Illinois 60612

Cell: 1-773-919-0088

Fax: 312 942 3666

Email: jai_raman@rush.edu

## Abstract

**Importance:** Understanding the causes and patient impacts of surgical adverse events will help improve systems and operational practices to avoid incidents in the future.

**Objective:** To determine the frequency, causes, and patient impact of adverse events in robotic procedures across different surgical specialties.

**Methods:** We analyzed the adverse events data related to robotic systems and instruments used in minimally invasive surgery, reported to the U.S. Food and Drug Administration (FDA) MAUDE database from January 2000 to December 2013. We determined the number of events reported per procedure and per surgical specialty, the most common types of device malfunctions and their impact on patients, and the causes for catastrophic events such as major complications, patient injuries, and deaths.

**Results:** During the study period, 144 deaths (1.4% of the 10,624 reports), 1,391 patient injuries (13.1%), and 8,061 device malfunctions (75.9%) were reported. The numbers of injury and death events per procedure have stayed relatively constant since 2007 (mean=83.4, 95% CI, 74.2–92.7). ***Surgical specialties,*** for which robots are extensively used, such as gynecology and urology, had lower number of injuries, deaths, and conversions per procedure than more complex surgeries, such as cardiothoracic and head and neck (106.3 vs. 232.9, Risk Ratio = 2.2, 95% CI, 1.9-2.6). ***Device and instrument malfunctions,*** such as falling of burnt/broken pieces of instruments into the patient (14.7%), electrical arcing of instruments (10.5%), unintended operation of instruments (8.6%), system errors (5%), and video/imaging problems (2.6%), constituted a major part of the reports. ***Device malfunctions impacted patients*** in terms of injuries or procedure interruptions. In 1,104 (10.4%) of the events, the procedure was interrupted to restart the system (3.1%), to convert the procedure to non-robotic techniques (7.3%), or to reschedule it to a later time (2.5%).

**Conclusions:** Despite widespread adoption of robotic systems for minimally invasive surgery, a non-negligible number of technical difficulties and complications are still being experienced during procedures. Adoption of advanced techniques in design and operation of robotic surgical systems may reduce these preventable incidents in the future.

## Introduction

The use of robotic systems for minimally invasive surgery has exponentially increased during the last decade. Between 2007 and 2013, over 1.74 million robotic procedures were performed in the U.S., of which over 1.5 million (86%) were performed in gynecology and urology, while the number of procedures in other surgical specialties altogether was less than 250,000 (14%)[1]. Several previous studies on the outcomes and rates of complications during robotic procedures in the areas of gynecology, urology, and general surgery have been done. Yet no comprehensive study of the safety and reliability of surgical robots has been performed.

Our study focuses on analysis of all the adverse events related to robotic surgical systems, collected by the FDA MAUDE database[2] during the 14-year period of 2000–2013. It covers the events experienced during the robotic procedures in six major surgical specialties: gynecology, urology, general, colorectal, cardiothoracic, and head and neck surgery. We analyzed the safety-related incidents, including deaths, injuries, and device malfunctions, to understand their causes and measure their impact on patients and on the progress of the surgery.

There have been several reports by different surgical institutions on occasional software-related, mechanical, and electrical failures of system components and instruments during robotic procedures[3-16]. A few studies analyzed the FDA MAUDE reports related to robotic surgical systems[17-23] (see Tables 1 and 2 in Appendix). However, most of the previous work targeted only two common robotic surgical specialties of gynecology and urology, or only analyzed small subsets or specific types of device failure modes (e.g., electro-cautery failures, electrosurgical injuries, instrument failures).

An important question is whether the evolution of the robotic systems with new technologies and features over the years has improved the safety of robotic systems and their effectiveness across different surgical specialties. Our goal is to use the knowledge gained from this analysis to provide insights on design of future surgical systems that by taking advantage of advanced safety mechanisms, improved human machine interfaces, and regulated operational practices can minimize the adverse impact on both the patients and surgical teams.

# Methods

## Data Sources

The Manufacturer and User Facility Device Experience ("MAUDE") database is a publicly available collection of *suspected medical device-related* adverse event reports, submitted by mandatory (user facilities, manufacturers, and distributors) and voluntary (health care professionals, patients, and customers) reporters to the FDA[2]. Manufacturers and the FDA regularly monitor these reports to detect and correct device-related safety issues in a timely manner. Each adverse event report contains information such as *Device Name*; *Manufacturer Name*; *Event Type* ("Malfunction," "Injury," "Death," or "Other"); *Event Date*; *Report Date*; and human-written *Event Description* and *Manufacturer Narrative* fields, which provide a short description of the incident, as well as any comments made or follow-up actions taken by the manufacturer to detect and address device problems[2].

While the MAUDE database, as a spontaneous reporting system, suffers from underreporting and inconsistencies[24,25,26], it provides valuable insights on real incidents that occurred during the robotic procedures and impacted patient safety. We treated the reported data on deaths, injuries, and device malfunctions provided by the MAUDE as a *sample set* to estimate the *lower bounds on prevalence of adverse events* and identify *examples* of their major causes and patient impacts (see eMethods for more details).

## Data Analysis Methods

We extracted all the reports related to the systems and instruments used in robotic surgery by searching for related keywords in the *Device Name* and *Manufacturer Name* fields of the MAUDE records posted between January 2000 and December 2013. In addition to the structured information that was directly available from the reports, we extracted further information from the unstructured human-written descriptions of events by natural language parsing of the *Event Description* and *Manufacturer Narrative* fields. We did so by creating several domain-specific dictionaries (e.g., for patient complications, surgery types, surgical instruments, and malfunction types) and pattern-matching rules as well as parts-of-speech (POS) and negation taggers to interpret the semantics of the event descriptions (Figure 1 in Appendix). The results generated by the automated analysis tools were manually reviewed for accuracy and validity. We extracted the following information:

- Patient injury (such as burns, cuts, or damage to organs) and death events that were reported under another *Event Type*, such as “Malfunction” or “Other”.
- Surgical specialty and type of robotic procedure during which the adverse events occurred.
- Major types of device or instrument malfunctions (e.g., falling of burnt/broken pieces of instruments into patients’ bodies or electrical arcing of instruments)
- Adverse events that caused an interruption in the progress of surgery, by leading the surgical team to troubleshoot technical problems (e.g., restarting the system), convert the procedure to non-robotic surgical approaches (such as laparoscopy or open surgery), or abort the procedure and reschedule it to a later time.

We compared the number of adverse events (in general) and injury/death events and procedure conversions (in particular) per 100,000 procedures across different surgical specialties. The rate of events was estimated by dividing the number of adverse events that occurred in each year (based on the *Event Date*) by the annual number of robotic procedures performed in the U.S. The total number of procedures per year was extracted from the device manufacturer’s reports[1,27] for 2004–2013 (see Figure 2 in Appendix). The annual number of procedures per surgical specialty was available only for gynecology, urology, and general surgery after 2007. So we estimated a combined annual number of procedures for cardiothoracic and head and neck surgery by assuming that the majority of the remaining procedures (other than genecology, urology, and general) were related to these specialties, as, according to the manufacturer reports, they are the only other specialties for which the robot has been used[1].

We assumed that the rate of underreporting for injury and death events are low and are independent from the type of surgery, because the device manufacturers are required and monitored by the FDA to report serious injury and death events to the MAUDE database. However, due to possible changes in the reporting rates during the years, the total number of events per procedure in the whole study period was compared across different surgical specialties. The 2-sided P values (< 0.05) and 95% confidence intervals were used to determine the statistical significance of the results.

To characterize the major causes to which injury and death events were attributed, we performed a manual review of event descriptions for all the reports made before 2013. The cumulative number of malfunctions per procedure was used to evaluate the trends in malfunction rates over 2004–2013.

## Results

We extracted a total of 10,624 events related to the robotic systems and instruments, reported over 2000–2013. About 98% of the events were reported by device manufacturers and distributors, and the rest (2%) were voluntary reports.

Data included 1,535 (14.4%) adverse events with significant negative patient impacts, including injuries (1,391 cases) and deaths (144 cases), and over 8,061 (75.9%) device malfunctions. For the rest of the events (1,028 cases), the *Event Type* information either was not available or was indicated as "Other." We identified 160 adverse events (1.5%) that included some kind of patient injuries but were reported as a "Malfunction" or "Other."

***Trends in Adverse Event Reports:*** Figure 1 shows the overall trends in the annual numbers of reports and the estimated rates of events per 100,000 procedures over 2004–2013:

- The absolute number of reports has significantly increased (about 32 times) since 2006, reaching 58 deaths, 938 patient injuries, and 4,124 malfunctions in 2013.
- While the annual average number of adverse events was about 550 per 100,000 procedures (95% confidence interval (CI), 410–700) between 2004 and 2011, in 2013 it peaked at about 1,000 events per 100,000 procedures.
- The numbers of injury and death events per procedure have stayed relatively constant since 2007 (mean=83.4, 95% CI, 74.2–92.7).

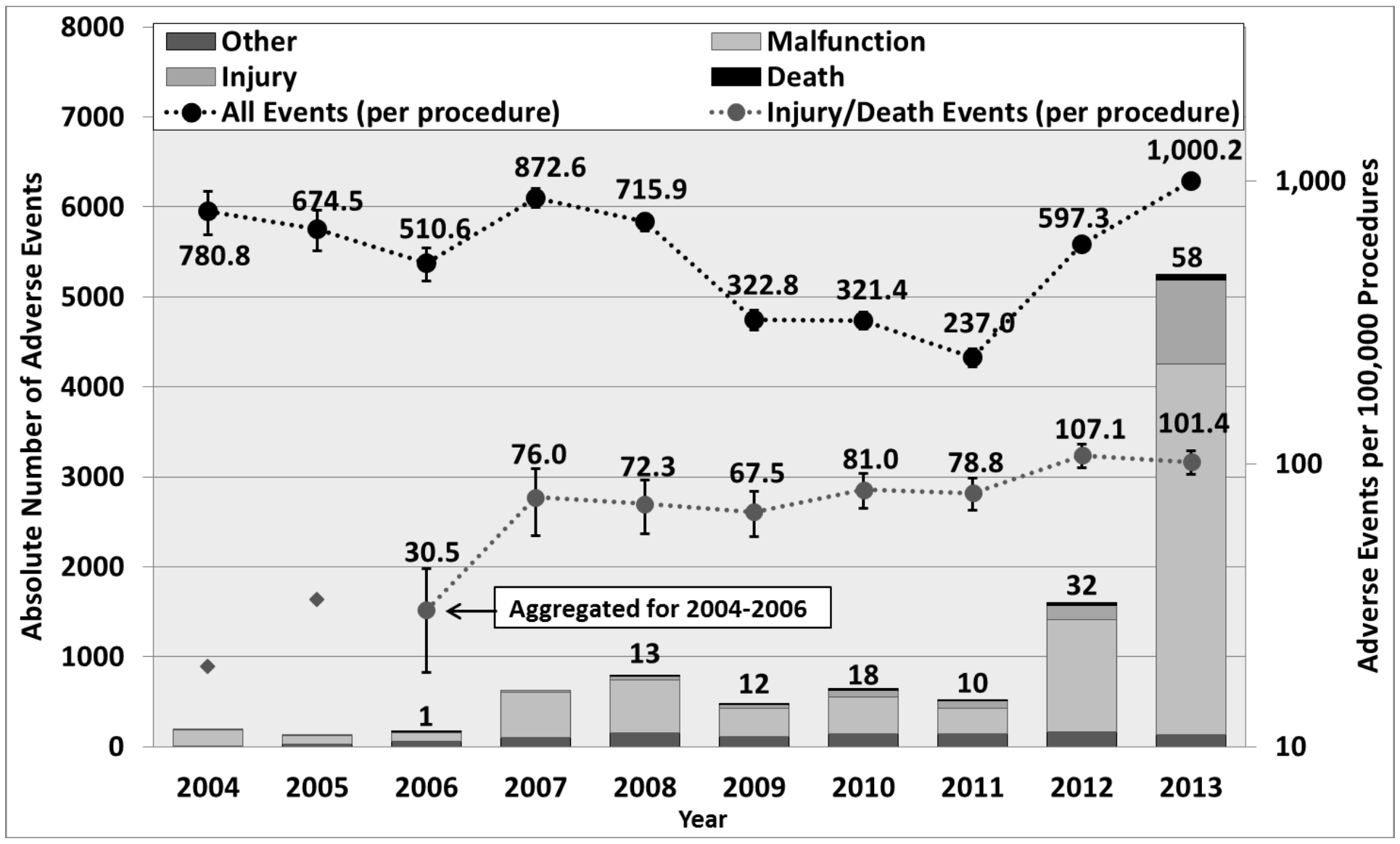

**Figure 1. Annual Numbers of Adverse Event Reports and Rates of Events per Procedure**

The left Y-axis corresponds to the bars showing the absolute numbers of adverse events (based on the year that reports were received by the FDA). The right Y-axis corresponds to the trend lines showing (in logarithmic scale) the annual number of adverse events per 100,000 procedures (based on the year the events occurred). Numbers on the bars indicate number of deaths reported per year. Error bars represent 95% confidence intervals for the proportion estimates. Because of the small number of injury and death events reported for 2004 and 2005, a combined rate was calculated for 2004–2006. Note that of all the events, 40 were reported as part of the articles or the legal disputes received by the manufacturing company.

***Adverse Events across Different Surgical Specialties***: Table 1 shows the numbers of adverse events reported in different surgical specialties and their impact on patients (injuries or deaths) and progress of surgery (procedure conversion or rescheduling). The last row shows examples of the most common types of procedures reported in each specialty.

- The majority of reports were related to gynecology (30.1%), urology (14.7%), and cardiothoracic (3.7%) surgeries, such as hysterectomy (2,331), prostatectomy (1,291), and thoracic (110) procedures, respectively.
- Cardiothoracic and head and neck surgeries involved a higher number of deaths per adverse event report (6.4% and 19.7%) than gynecology and urology (1.4 and 1.9%).
- The highest number of procedure conversions per adverse event was for cardiothoracic (16.8%) and urology (13.5%), and the highest rates of procedure rescheduling were for urology (9.5%), general (3.0%), and cardiothoracic (2.8%) surgeries.

**Table 1. Adverse events in different surgical specialties: Deaths, injuries, malfunctions, procedure conversion or rescheduling, common types of surgery**

| | No. (%) [95% Confidence Interval] | | | | | | |
|---|---|---|---|---|---|---|---|
| | **Gynecology** | **Urology** | **Cardiothoracic** | **Head & Neck** | **Colorectal** | **General** | **N/A** |
| **Overall** [a] | 3,194 (30.1) [29.2–31.0] | 1,565 (14.7) [14.0–15.4] | 393 (3.7) [3.3–4.1] | 71 (0.7) [0.5–0.9] | 301 (2.8) [2.5–3.1] | 197 (1.9) [1.6–2.2] | 4,903 (46.2) [45.3–47.1] |
| **Event Type** [b] | | | | | | | |
| Death | 46 (1.4) [1.0–1.8] | 30 (1.9) [1.2–2.6] | 25 (6.4) [4.0–8.8] | 14 (19.7) [10.4–29.0] | 11 (3.7) [1.6–5.8] | 11 (5.6) [2.4–8.8] | 7 (0.1) [0.0–0.2] |
| Injury | 818 (25.6) [24.1–27.1] | 272 (17.4) [15.5–19.3] | 64 (16.3) [12.6–20.0] | 14 (19.7) [10.4–29.0] | 58 (19.3) [14.8–23.8] | 56 (28.4) [22.1–34.7] | 109 (2.2) [1.8–2.6] |
| Malfunction | 2,103 (65.8) [64.2–67.4] | 902 (57.6) [55.2–60.0] | 226 (57.5) [52.6–62.4] | 35 (49.3) [37.7–60.9] | 209 (69.4) [64.2–74.6] | 110 (57.8) [48.9–62.7] | 4,476 (91.3) [90.5–92.1] |
| Other | 227 (7.1) [6.2–8.0] | 361 (23.1) [21.0–25.2] | 78 (19.8) [15.9–23.8] | 8 (11.3) [3.9–18.7] | 23 (7.6) [4.6–10.6] | 20 (10.2) [6.0–14.4] | 311 (6.3) [5.6–7.0] |
| **Conversion** | 236 (7.4) [6.5–8.3] | 212 (13.5) [11.8–15.2] | 66 (16.8) [13.1–20.5] | 6 (8.5) [2.0–15.0] | 29 (9.6) [6.3–12.9] | 14 (7.1) [3.5–10.7] | 217 (4.4) [3.8–5.0] |
| **Rescheduling** | 26 (0.8) [0.5–1.1] | 148 (9.5) [8.1–10.9] | 11 (2.8) [1.2–4.4] | 1 (1.4) [0–4.1] | 1 (0.3) [0–1.0] | 6 (3.0) [0.6–5.4] | 77 (1.6) [1.3–1.9] |
| Common Surgery Types | Hysterectomy (2,331) | Prostatectomy (1,291) | Thoracic (110) | Thyroidectomy (19) | Cholecyst-ectomy (118) | Hernia repair (37) | |
| | Myomectomy (328) | Nephrectomy (138) | Lobectomy (67) | Tongue base resection (19) | Colectomy (61) | Nissen fundoplication (34) | |
| | Sacrocolpopexy (170) | Pyeloplasty (31) | Mitral valve repair (54) | Transoral robotic (18) | Low anterior resection (44) | Gastric bypass (28) | |
| | Oophorectomy (120) | Cystectomy (48) | Coronary artery bypass (23) | | Colon resection (25) | Gastrectomy (15) | |

[a] Percentages are over all the adverse event reports (n = 10,624).
[b] Percentages are over the total adverse events reported for a surgical specialty.

The higher percentage of adverse events reported in gynecology and urology could be due to the higher number of these procedures performed.

Of all the reports, only 5,721 (53.8%) indicated the class and type of surgery involved. However, the majority of reports with missing information on the type of surgery were related to device malfunctions and "Other" events (97.6%). In order to compare the rate of adverse events across different specialties, we focused only on reports related to injuries, deaths, and procedure conversions. For the majority of these events (92.2% of injury reports, 95.1% of deaths, and 72.2% of procedure conversions), the surgery type information was available and the rest (with 'N/A' surgical specialty) were removed from our analysis. In order to estimate the rate of events per procedure, we regrouped the events into four major categories of "Gynecology," "Urology," "General," and "Cardiothoracic and Head and Neck," according to the manufacturer's reports[1,27]. The "General" category includes both colorectal and general specialties.

As shown in Table 2, for cardiothoracic and head and neck surgery, the rates of injuries, deaths, and procedure conversions have been significantly higher than other specialties. During 2007-2013, the estimated rate of deaths have been 52.2 per 100,000 procedures for cardiothoracic and head and neck specialties vs. 5.7 in gynecology, urology, and general surgeries (RR = 9.23, 95% CI, 6.35–13.40, P < 0.0001). Also, the rate of injuries and procedure conversions in these specialties have been 91.0 and 89.7 per 100,000 procedures vs. 71.5 (RR = 1.27, 95% CI, 0.99–1.63, P < 0.052) and 29.2 (RR = 3.07, 95% CI, 2.38–3.97, P < 0.0001) in the other surgical categories.

**Table 2. Comparsion of adverse events rates in different surgical specialities (2007 - 2013)**

| | No. (rate per 100,000 procedures) [a] [95% CI] | | Cardiothoracic and Head and Neck vs. Gynecology, Urology, and General | |
|---|---|---|---|---|
| | **Gynecology, Urology, General** | **Cardiothoracic, Head and Neck, Other** | **Relative Risk (95% CI)[b]** | ***P* Value** |
| Total Procedures | 1,661,891 | 74,709 | | |
| Total Adverse Events | 5,209 | 447 | | |
| Event Type | | | | |
| Death | 94 (5.7) | 39 (52.2) | 9.23 (6.35–13.40)1 | < 0.0001 |
| Injury | 1188 (71.5) | 68 (91.0) | 1.27 (0.99–1.63) | < 0.052 |
| Conversion | 485 (29.2) | 67 (89.7) | 3.07 (2.38–3.97) | < 0.0001 |
| Rescheduling | 180 (10.8) | 12 (16.1) | 1.48 (0.83–2.66) | < 0.19 [c] |

[a] Percentages are over total number of procedures in each column.
[b] Assuming that the level of underreporting across different surgical specialties is similar.
[c] Not statistically significant because of the small number of samples (12) in the cardiothoracic and head and neck surgery.

***Device and Instrument Malfunctions:*** We identified five major categories of device and instrument malfunctions experienced during procedures that impacted the patients, either by causing injuries and complications or by interrupting the progress of surgery and/or prolonging procedure times. Table 2 shows the numbers of events in each category, the event types as indicated by reporters, and the actions taken by the surgical team to resolve the problems. The *Other* category includes the malfunctions that could not be classified in any of the classes.

- **System errors and video/imaging problems** contributed to 787 (7.4%) of the adverse events and were the major contributors to the system resets (274 cases, 82% of all system resets), conversion of the procedures to a non-robotic approach (462 cases, 59.2% of all conversions), and aborting/rescheduling of the procedures (221 cases, 81.8% of all cases).
- **Falling of the broken/burnt pieces into the patient's body** constituted about 1,557 (14.7%) of the adverse events. In almost all these cases, the procedure was interrupted, and the surgical team spent some time searching for the missing pieces and retrieving them from the patient (in 119 cases, a patient injury, and in one case a death, was reported).
- **Electrical arcing, sparking, or charring of instruments** and burns or holes developed in the tip cover accessories constituted 1,111 reports (10.5% of the events), leading to nearly 193 injuries, such as burning of tissues.
- **Unintended operation of instruments**, such as uncontrolled movements and spontaneous powering on/off, happened in 1,078 of the adverse events (10.1%), including 52 injuries and 2 deaths.

In total, 5,054 reports (47.6%) were related to breakage of different parts of the system and instruments. Cable, wire, or tube breakages are example causes of imaging problems at the surgeon's console or unintended instrument operations.

**Table 3. Major categories of malfunctions**

| Malfunction Category [a] | Description | No. of Reports (% of all) | | | | | Surgical Team Actions (% of malfunction category) | | |
|---|---|---|---|---|---|---|---|---|---|
| | | Total | Event Type [b] | | | | System Reset | Procedure Converted | Procedure Rescheduled |
| | | | M | IN | D | O | | | |
| **System Errors [c]** | - System error codes and faults<br>- System transferred into a recoverable or non-recoverable safety state | 536 (5.0%) | 44 | 23 | 1 | 468 | 231 (43.1%) | 330 (61.6%) | 133 (24.8%) |
| **Video/ Imaging Problems** | - Loss of video<br>- Display of blurry images at surgeon's console or assistant's touchscreen | 275 (2.6%) | 21 | 18 | 0 | 236 | 53 (19.3%) | 145 (52.7%) | 94 (34.2%) |
| **Broken Pieces Falling Into Patients** | - Burnt/broken parts and components<br>- Fell into surgical field or body cavity<br>- Required additional procedure time to be found/removed from the patient | 1,557 (14.7%) | 1,396 | 119 | 1 | 41 | 3 (0.2%) | 38 (2.4%) | 5 (0.3%) |
| **Broken Tip Covers/ Elec. Arcing** | - Tears, burns, splits, holes on tip cover<br>- Electrical arcing, sparking, charring | 1,111 (10.5%) | 900 | 193 | 0 | 18 | 2 (0.2%) | 18 (1.6%) | 0 (0.0%) |
| **Unintended Instrument Operation** | - Unintended or unstoppable movements started without the surgeon's command<br>- Instruments not working, open/closed<br>- Instruments not recognized by system | 1,078 (10.1%) | 919 | 52 | 2 | 105 | 31 (2.9%) | 93 (8.6%) | 21 (1.9%) |
| **Other** | - Issues with electrosurgical units, power supplies/cords, patient-side manipulators, etc.<br>- Other events reported as "Malfunction" | 5,026 (47.3%) | 4,962 | 31 | 1 | 32 | 20 (0.4%) | 41 (0.8%) | 10 (0.2%) |
| **Total (% of all)** | - All malfunctions [d] | 9,382 (88.3%) | 8,061 | 444 | 9 | 868 | 305 (3.2%) | 631 (6.7%) | 246 (2.6%) |
| | **All Adverse Events** | 10,624 (100%) | 8,061 | 1,391 | 144 | 1,028 | 334 (3.1%) | 780 (7.3%) | 270 (2.5%) |

| | System Errors | Video/Imaging Problems | Broken Pieces Fell into Patients | Burns/Holes in Tip Covers/Elec. Arcing | Unintended Instrument Operation | Other |
|---|---|---|---|---|---|---|
| **Broken Instruments (n = 5,054)** | 33 (6.2%) | 27 (9.8%) | 1,166 (74.9%) | 275 (24.8%) | 400 (37.1%) | 3,201 (63.7%) |
| | clamps | camera, cables | instrument pieces, tips, cautery hooks | tips, wires, insulation | cables, tubes, wires | |

[a] The malfunction categories and actions taken by the surgical team are not mutually exclusive, and in many cases two or three different malfunctions or two actions were reported in a single event. Figure 3 in Appendix uses Venn diagrams to depict the intersections among different malfunction categories and actions taken by the surgical team.

[b] Event types as indicated by reporters: Malfunction (M), Injury (IN), Death (D), and Other (O).

[c] Table 3 in Appendix lists the descriptions and frequencies of the most common system error codes extracted from the reports.

[d] In 1,019 of cases (10.9% of all the malfunctions), the device or instrument malfunction was detected prior to start of the procedure, of which in 20 cases the procedure was rescheduled to a later time and in 2 cases it was converted to a non-robotic approach.

Figure 3 shows the cumulative rates of malfunctions per procedure over 2004–2013. Overall, the malfunction rates decreased after 2006, but the rate of cases with arcing instruments and broken instruments followed a relatively constant trend. The sudden increase in the rate of broken instruments after the middle of 2012 could be related to changes made to the adverse event reporting practices by the manufacturer in 2012 (mostly related to instrument cable breaks)[30], as well as increased reporting of adverse events after concerns about the safety of robotic surgery were raised by the FDA[31,32] and public media in early 2013[33,34,35].

In total, 9,382 reports were about technical problems, including 1,104 cases (10.4% of all the adverse events) in which the procedure was interrupted and additional time was spent on troubleshooting the errors, resetting the system, and/or converting the procedure to a traditional technique, or rescheduling the procedure to a later time.

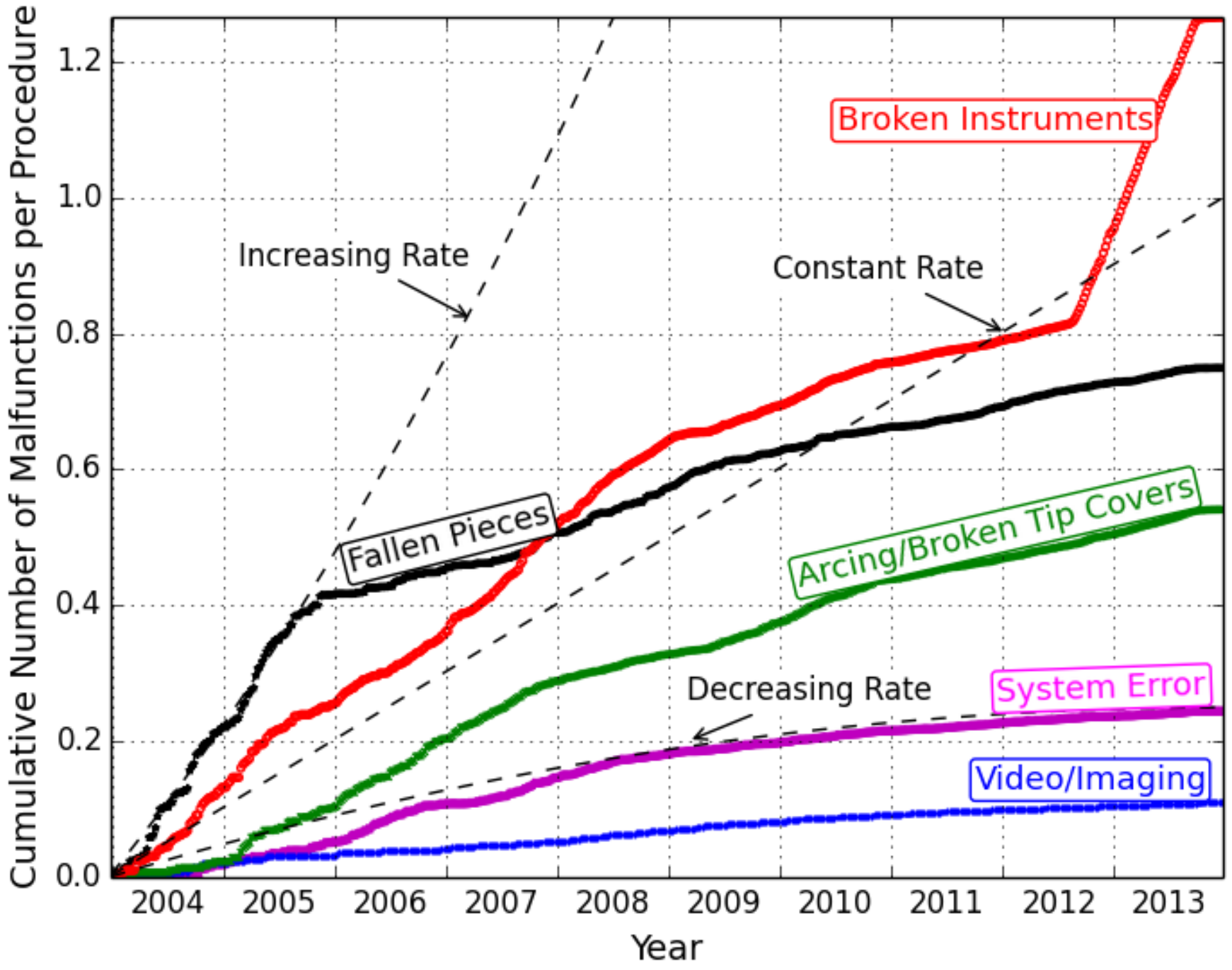

**Figure 2. Cumulative rates of malfunctions per procedure**

The rates of malfunctions per procedure were obtained for each week (see Figure 2 in Appendix for more details).

***Injury and Death Causes:*** A manual review of a sample set of injury and death reports (from 2000–2012) was conducted. This allowed us to classify the causes indicated by reporters into three main categories: inherent risks associated with surgery, technical issues with the robot, and mistakes made by the surgical team. For the majority of death events, little or no information was provided in the reports. About 33.7% of the death events were related to inherent risks or complications during surgery, and 7% were attributed to operator mistakes. About 62% of the injury events involved device malfunctions (see Table 2), and the rest were related to operator errors (7.1%), improper positioning of patient or port incisions (6.3%), inherent risks of surgery (3.9%), or problems with grounding the equipment (1.5%) (see Tables 4 and 5 in Appendix).

## Discussion

Our analysis shows an increasing number of adverse events related to the robotic surgical systems being reported. As cautioned by the FDA[2,31], the number of MAUDE reports may not be used to evaluate the changes in rates of events over time, because the increased reporting of events may be due to different factors, e.g., the increasing use of surgical systems[1], changes in the manufacturers' reporting practices[30], and/or better awareness and increased publicity resulting from product recalls, media coverage, and litigation[31]. Therefore, we measured the prevalence of adverse events by estimating the number of events reported per procedure. We found that despite a relatively high number of reports, the vast majority of procedures were successful and did not involve any problems and the number of injury/death events per procedure have stayed constant since 2007.

However, our analysis shows that estimated number of events per procedure in complex surgical areas, such as cardiothoracic and head and neck surgery were significantly higher than gynecology, urology, and general surgeries. Although not all the reported injuries and deaths were due to device problems, and the procedure conversions, of themselves, cannot be considered adverse events[41,42], the estimated numbers of injury/death events and conversions per procedure are used as a metric to measure the difficulty experienced in different surgical specialties. The best that we can tell from the available data is that the higher number of injury, death, and conversion *per adverse event*, in cardiothoracic and head and neck surgeries, could be indirectly explained by the higher complexity of the procedures, less frequent use of robotic devices, and less robotic expertise in these fields. Although the use of robotic technology has rapidly grown in urology and gynecology for prostatectomy and hysterectomy, it has been slow to percolate into more complex areas, such as cardiothoracic and head and neck surgery. The limitations of the robotic user interface[36], long procedure times[37], learning curve[38,39], and higher costs[40] are some factors that may have contributed to the lower utilization of the robotic approach in more complex surgical procedures. For example, only a select type of robotic cardiac procedures are reported to have been successfully performed using the robots, such as mitral valve repair and internal mammary artery harvest[43,44,45]. The recent experiences of highly competent robotic teams that performed multi-vessel coronary artery bypass grafting (CABG) showed that the robotic approach may be associated with higher mortality and morbidity rates compared to open surgery[46].

In practice, the use of the robotic platform involves the interface of a sophisticated machine (see Table 6 in Appendix) with surgical teams, in an area of patient care that is safety-critical. From a technology perspective, some of the reported events could be prevented by employing substantially improved safety practices and controls in the design and operation of surgical systems. Some examples include:

- New safety engines for monitoring of procedures (including surgeon, patient, and device status) and providing comprehensive feedback to surgical team on upcoming events and troubleshooting procedures to prevent long procedure interruptions.
- Providing real-time feedback to the surgeon on the safe surgical paths that can be taken[47], by computing 3D models of the organs under surgery and surrounding critical tissues and vessels, as well as surgeon-specific modeling and monitoring of robotic surgical motions[48], to minimize the risk of approaching dangerous limits and inadvertent patient injuries.
- Improved human-machine interfaces and surgical simulators that train surgical teams for handling technical problems and assess their actions in real-time during the surgery.

## Limitations

The results of our study come with the caveats that inherent risks exist in all surgical procedures (more so in complex procedures) and that the MAUDE database suffers from underreporting and inconsistencies. Thus, the estimated number of adverse events per procedure are likely to be lower than the actual numbers in robotic surgery. Further, the lack of detailed information in the reports makes it difficult to determine the exact causes and circumstances underlying the events. Therefore, the sensitivity of adverse event trends to changes in reporting mechanisms, surgical team expertise, and inherent risks of surgery could not be assessed here.

## Conclusions

While the robotic surgical systems have been successfully adopted in many different specialties, this study demonstrates several important findings: (1) the overall numbers of injury and death events per procedure have stayed relatively constant over the years, (2) the probability of events in complex surgical specialties of cardiothoracic and head and neck surgery has been higher than other specialties, (3) device and instrument malfunctions have affected thousands of patients and surgical teams by causing complications and prolonged procedure times.

As the surgical systems continue to evolve with new technologies, uniform standards for surgical team training, advanced human machine interfaces, improved accident investigation and reporting mechanisms, and safety-based design techniques should be developed to reduce incident rates in the future.

# Appendix

## Underreporting

The underreporting in data collection is a fairly common problem in social sciences, public health, criminology, and microeconomics. It occurs when the counting of some event of interest is for some reason incomplete or there are errors in recording the outcomes. Examples are unemployment data, infectious or chronic disease data (e.g. HIV or diabetes), crimes with an aspect of shame (e.g. sexuality and domestic violence), error counts in a production processes or software engineering, and traffic accidents with minor damage [1]. An estimated prevalence of events based on the incomplete counts is likely to be smaller than the true proportion of events in the population. Several inference techniques based on binomial, beta-binomial, and regression models have been proposed for estimating the actual count values [2]. However, in all those techniques the reporting probability (underreporting rate) is assumed to be a constant parameter over time that is estimated based on the sample counts.

A very similar problem exists in preliminary or pilot clinical investigations, epidemiological surveys, and longitude studies where the objective is to estimate any possible clinical effect of a treatment or prevalence of a particular disease in a population of patients, but the prevalence of events can only be estimated by selecting a sample of patients from the population [3].

In all these situations, the prevalence of the events are estimated based on a random sample of events from the population, under the assumption that the sample set contains the same characteristics and distributions of the actual population, including those of the underreported and missing cases.

Furthermore, it is often required to perform a sample-size calculation based on confidence intervals in order to provide a precise estimate with a large margin of certainty and to make sure that the estimated proportion is close to the actual proportion with a high probability [3]. Confidence intervals for the proportions estimated based on samples from large populations and finite populations can be calculated by using the normal approximation to the binomial distribution as follows:

For large populations:

$$p \pm z_{1-\alpha/2}\sqrt{\frac{p(1-p)}{N}}$$

For finite populations:

$$p \pm z_{1-\alpha/2}\sqrt{\left[\frac{p(1-p)}{N}.\frac{N_{Population-N}}{N_{Population}}\right]}$$

where $N$ is the size of sample, $p = \frac{r}{N}$ is the estimate of the proportion of events of interests in the sample and $N_{Population}$ is the size of population in case of finite populations [3].

In this study, we estimated the prevalence of adverse events by making sure that we have a significantly large enough number of samples to provide confident estimates. Our estimations are obtained under the assumption that the characteristics and distributions of the observed events are not significantly different from those in the actual population and would not significantly change after including the underreported cases. We are currently investigating the extension of the proposed inference techniques in [1][2] to estimate the actual number of adverse events with considering a variable reporting probability over time.

## Figure 1. Data extraction and analysis flow from the FDA MAUDE database

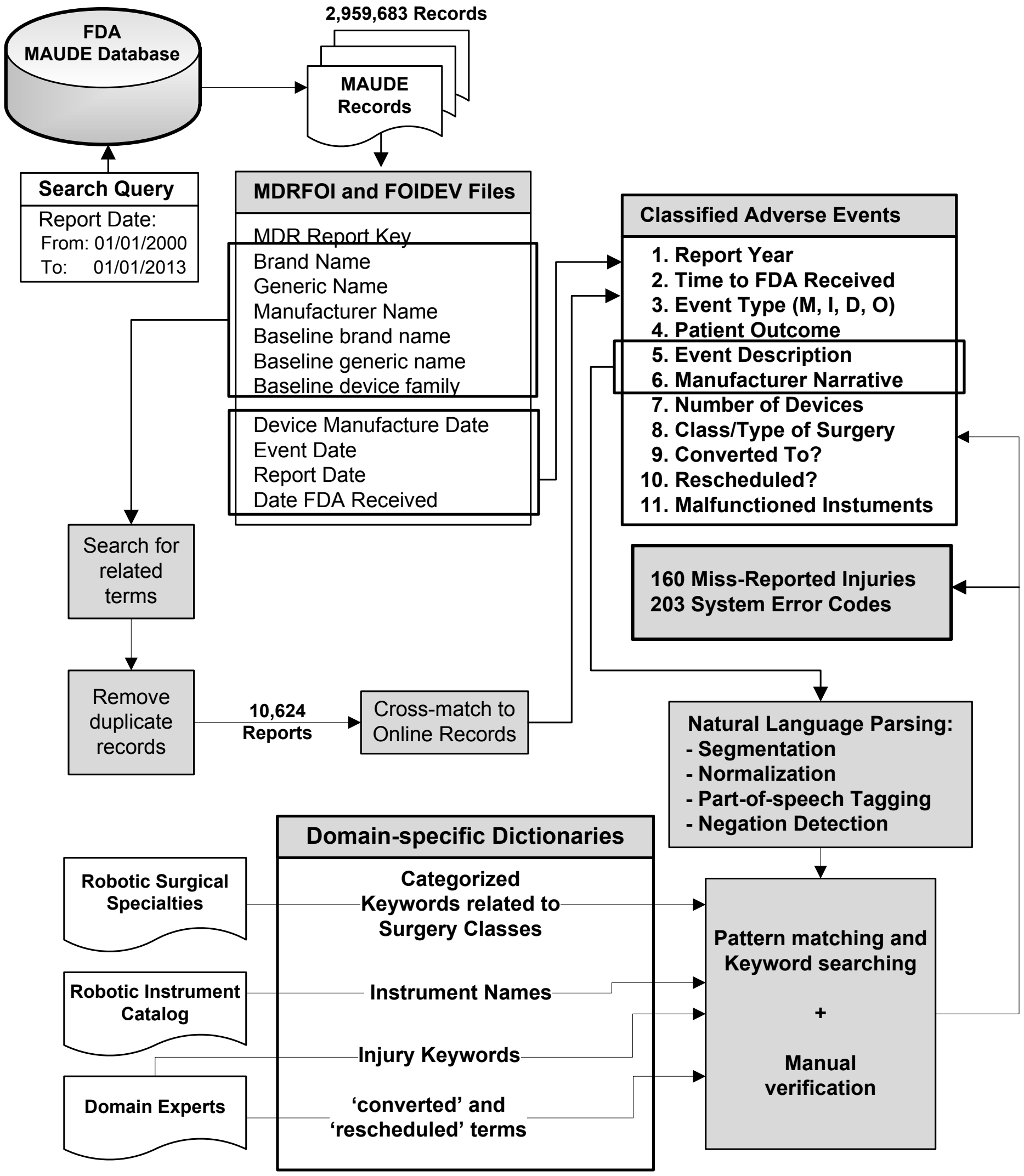

**Figure 2. Estimated numbers of procedures performed during 2004–2013**

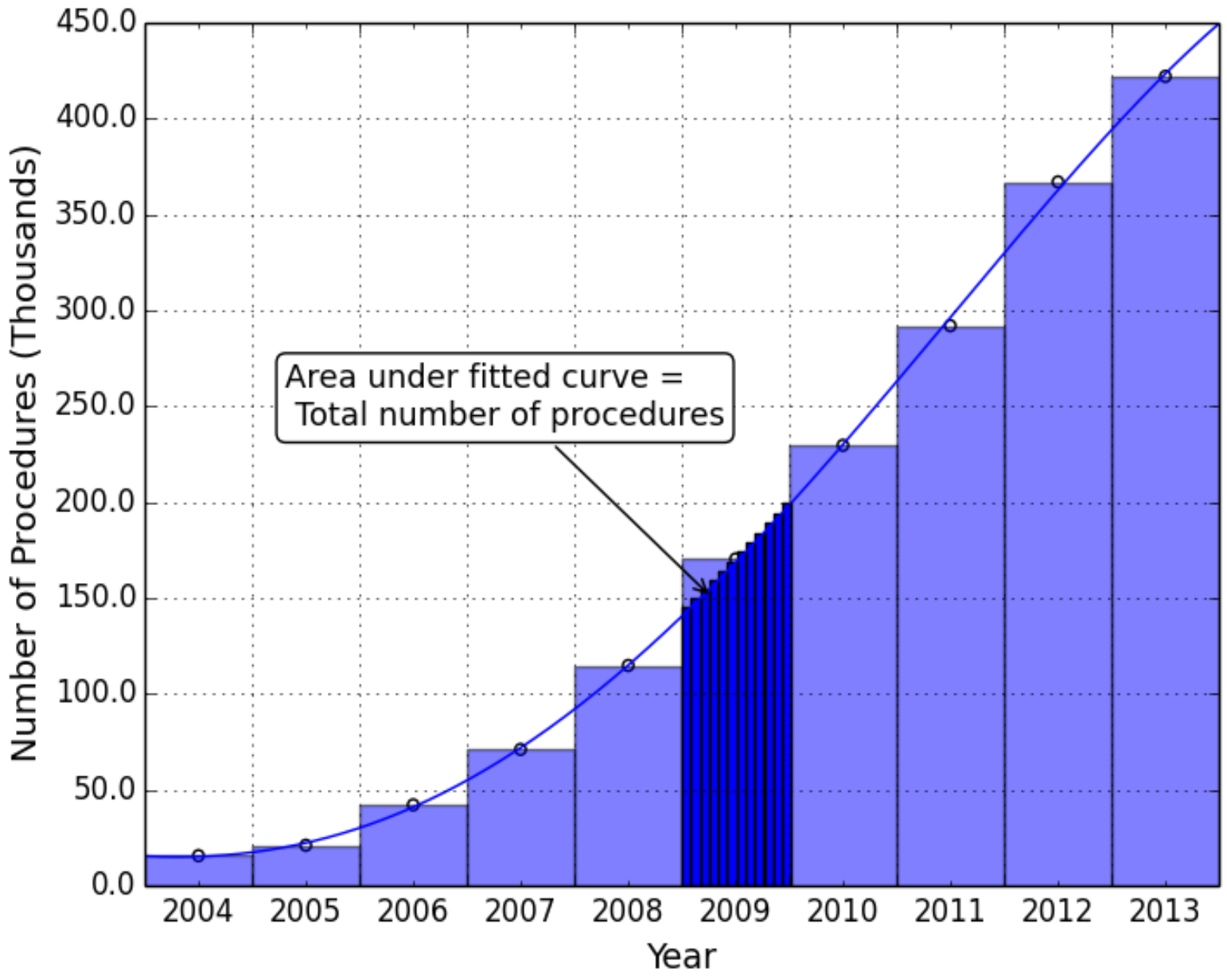

The annual numbers of procedures performed in the U.S. for 2010–2013 were extracted from the annual reports of the manufacturer[1]. For 2004–2009, we estimated the numbers of procedures by measuring the graphs in the company's investor presentations[27]. Whenever the estimated numbers from two different sources did not match or the data were available only for the total worldwide procedures, we chose the maximum number of procedures for that year in order to achieve a lower bound on the likelihood of events.

We estimated the number of procedures per week from annual number of procedures by fitting a 4-degree polynomial curve ($R^2 = 0.999$) to the bar graph of annual procedures and calculating the area under the fitted curve for every week.

**Figure 3. (a) Intersections among different malfunction categories, (b) Intersections among system resets, converted, and rescheduled cases**

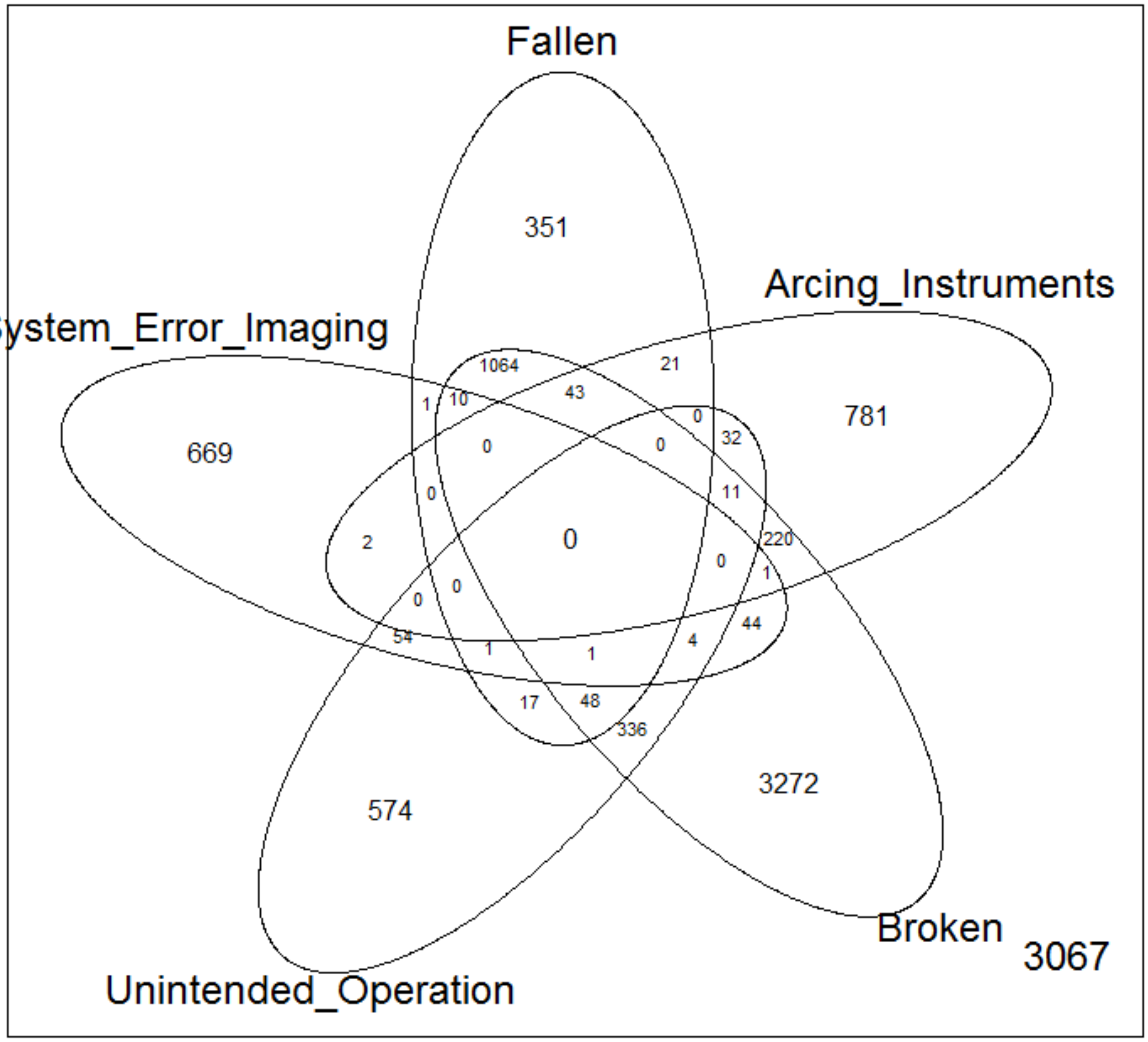

(A total of 3,067 adverse event reports were not classified by MedSafe in any of the malfunction categories.)

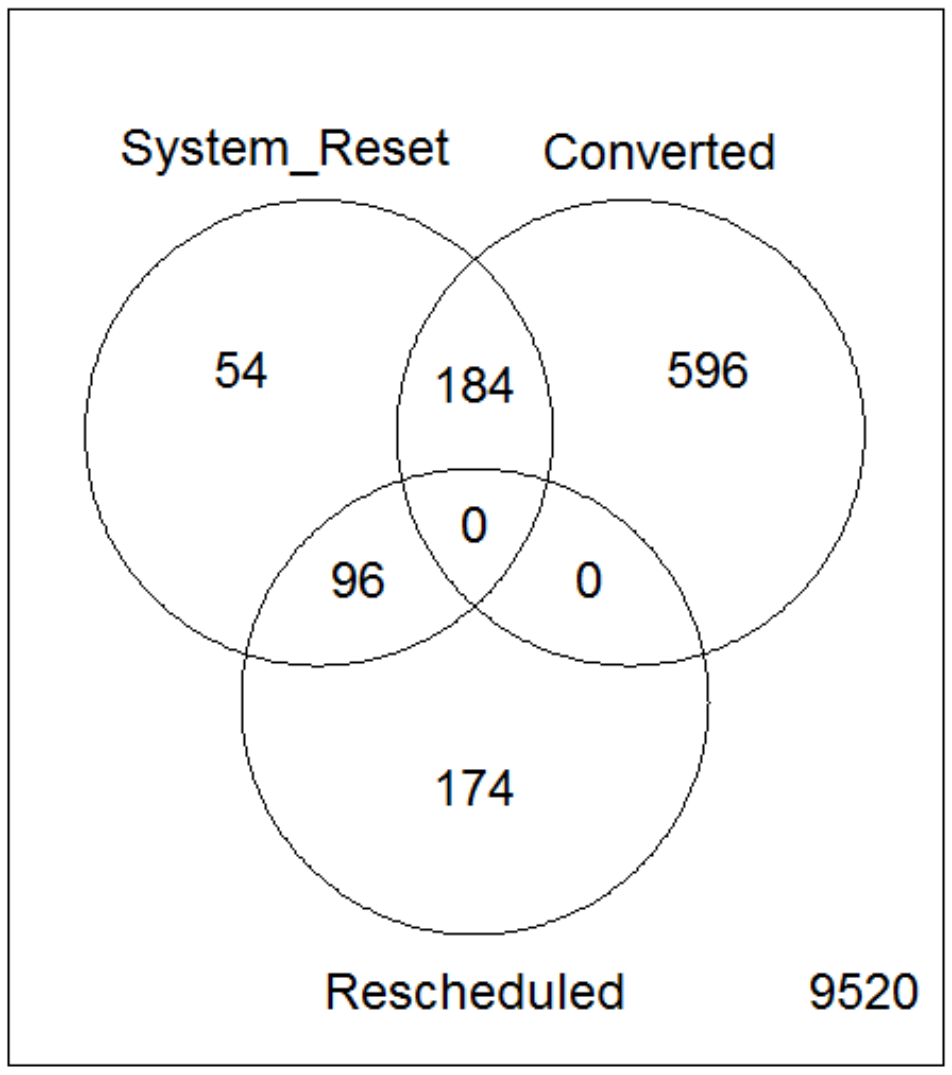

(For 9,520 adverse events, no system resets, conversions, or reschedulings were reported.)

## Table 1. Summary of related work on failures of robotic surgical systems

| Ref. No. (Year) | Surgery Types | Medical Institute | No. of Cases | Total Number of Failures (Failure Rate) Types of Malfunctions | Converted | Rescheduled |
|---|---|---|---|---|---|---|
| Eichel[3] (2005) | Urologic | UC Irvine | 200 | Total = 5 (2.5%)<br>Software (4), Mechanical (1) | N/A | N/A |
| Kozlowski[4] (2006) | Radical Prostatectomy (RLRP) | Virginia Mason Medical Center (VMMC) | 130 | Total = 6 (4.6%)<br>Setup joint (2), Software incompatible (1), Robotic arm (1), Power-off (1), Monitor loss (1) | Laparoscopic (1)<br>Open (1) | 1 |
| Borden[5] (2007) | Laparoscopic Prostatectomy | Virginia Mason Medical Center (VMMC) | 350 | Total = 9 (2.6%)<br>Setup joint (2), Robotic arm (1), Camera (1), Power error (1), Console metal break (1), Software incompatible (1), Monitor loss (1) | Laparoscopic/ Open (3) | 6 |
| Zorn[6] (2007) | Radical Prostatectomy (RLRP) | University of Chicago Pritzker School of Medicine (2003–2006) | 800 | Total = 7<br>(Recover. = 0.21%, Non-Recover. =.05%)<br>Power-up failure (1), Optical malfunction (3), Surgeon handicap (3), Robotic arm (1), Camera (2) | Completed (3) | 4 |
| Fischer[7] (2008) | Radical Prostatectomy | Klinik Hirslanden, Zurich, Switzerland | 210 | Total = 2 (1%)<br>Robotic arm (2) | Conventional Laparoscopic | N/A |
| Lavery[8] (2008) | Radical Laparoscopic Prostatectomy (RALP) | 11 Institutions<br>700 Surgeons | 8,240 | 34 critical failures (0.4%)<br>Robotic arm (14), Optical system (14), Masters (4), Power supply/circuit (6), Unknown error (3) | Laparoscopic (2)<br>Open (8) | 24 |
| Ham[9] (2009) | Radical Laparoscopic Prostatectomy | Yonsei University College of Medicine, Korea | 1 | Case report of Surgeon's console failure | Delayed 15 min | |
| Kim[10] (2009) | Urology, General Surgery, Obstetrics and Gynecology, Thoracic Surgery, Cardiac Surgery, Otorhinolaryngology | Yonsei University College of Medicine, Korea<br>(2005–2008) | 1,797 | Total = 43 (2.4%)<br>**Robot failures (24):** On/off failure (1), Console malfunction (5), Robotic arm (6), Optic system (2), System error (10)<br>**Instrument failures (19):** Shaft injuries (9), Wire cutting (2), Unnatural motion (2), Instrument tip (2), Limitation in motion (1) | Laparoscopic/ Open (3) | N/A |
| Kaushik[11] (2010) | Robot-assisted Radical Prostatectomy (RARP) | Survey of 176 Surgeons from 4 Countries | N/A | Total failures = 260<br>Robotic arm (38%), Camera (17.6%), Setup joint (13.8%), Power error (8.8%), Ocular monitor loss (8%), Instruments (7.6%), Console handpiece break (3%), Software (1.9%), Backup battery (0.3%), Instrument identification (0.3%) | Open (18.8%), Laparoscopic (15%), Another robot, with one fewer robotic arm (8.7%) | 57.5% |
| Finan[12] (2010) | Gynecologic Oncology | Mitchell Cancer Institute, University of South Alabama (2006–2008) | 137 | Total = 11 (8.02%)<br>Robotic arm (2), Light or camera cord (2), Maylard bipolar (1), Power failure (1), Port problem (1), Others (3) | Delayed 25 min. | N/A |
| Mues[13] (2011) | Urology, Gynecology, Cardiothoracic, General surgery, Otolaryngology, Neurosurgery | Ohio State University Medical Center, James Cancer Hospital (2008–2009) | 454 | Tip cover failures = 12 (2.6%)<br>Significant patient complications (25%) | Repaired at the time of surgery | N/A |
| Agcaoglu[14] (2012) | General Surgery | Cleveland Clinic | 223 | Total = 10 (4.5%)<br>Robotic instrument (4), Optical system (3), Robotic arms (2), Robotic console (1) | Open surgery (6) | N/A |
| Chen[15] (2012) | Urological Surgery | Veterans General Hospital, Taiwan (2005–2011) | 400 | Total = 14 (3.5%)<br>Robotic arm/joint (11), Optical system (1), Power system (1), Endoscopic instrument (1), Software incomp. (1) | Completed (10), Laparoscopy (3) | 1 |
| Buchs[16] (2014) | General Surgery | A Teaching Institution (2006-2012) | 526 | Total = 18 (3.4%)<br>Robotic instruments (9), Robotic arms (4), Surgical console (3), Optical system (2) | Laparoscopic (1) (0.2%) | N/A |

**Table 2. Summary of related work on analysis of MAUDE data of robotic surgical systems**

| |
|---|
| Murphy et al.[17] identified 38 system failures and 78 adverse events related to the da Vinci robotic system, reported from 2006 to 2007; most of them were related to broken instrument tips or failure of electrocautery elements. |
| Andonian et al.[18] found an estimated failure rate of 0.38% for robotic-assisted laparoscopic surgeries by reviewing 189 adverse events related to the ZEUS and da Vinci surgical robotic systems, reported to the MAUDE database between the years 2000 and 2007. |
| Lucas et al.[19] compared the rates of adverse events for two different models of da Vinci surgical systems (dVs and dV) during the period of 2003−2009 and showed that both device malfunctions and open conversions were reduced by increased robotic experience and newer surgical systems. |
| Fuller et al.[20] reviewed 605 adverse events involving the da Vinci system during 2001–2011, and identified 24 (3.9%) reports related to electrosurgical injuries (ESI) that occurred during gynecological and prostatectomy procedures. |
| Friedman et al.[21] analyzed the da Vinci robotic system instrument failures reported to the MAUDE database in 2009 and 2010. They found a total of 565 instrument failures, of which the majority were related to the instrument wrist or tip (285), 174 were related to cautery problems, 76 were shaft failures, and the rest were cable and control housing failures (36). |
| Gupta et al.[22] reviewed a total of 741 adverse events reported to the MAUDE database in 2009 and 2010. They found that 43.5% of the events were related to the use of energy instruments, that 30.97% were associated with the surgical systems and instruments, and that the severity of events correlated with the type of surgery and the type of device used. |
| Finally, Manoucheri et al.[23] evaluated the adverse events reported during robotic gynecologic procedures and found that the majority of reported injuries (65%) were not directly related to the surgical system; 21% were related to operator error; and 14% were due to technical system failures. |

## Table 3. Most frequent system error codes

| System Error Code | Description | Type of Safe State that System Transits To | No. of Adverse Events |
|---|---|---|---|
| #20008 | The angular position of one or more robotic joint's on the specified manipulator, as measured by the joint's primary control sensor (encoder) and the secondary sensor (potentiometer), were out of specified tolerance for agreement. | Recoverable | 62 |
| #23008 | | | 42 |
| #20013 | | | 34 |
| #23013 | | | 20 |
| #21008 | | | 18 |
| #21013 | | | 17 |
| #23002 | | | 8 |
| #20009 | | | 7 |
| #22003 | | | 5 |
| #212 | A voltage-tracking fault reported by the digital signal processor (dsp) when the actual voltage to drive current through the motors deviates from the expected voltage by a specified amount. | Non-recoverable | 31 |
| #23 | Hardware wheel "wdog" has tripped on one of the digital communication links in the system (due to an excessive number of retries on hardware message packets). This means that the system cannot reliably communicate over that digital link and therefore cannot continue normal operation.<br>Communication faults in the low-voltage differential signal carrying information about the patient side manipulator.<br>Communication faults between two system components. | N/A | 28 |
| #1 | A power supply voltage was out of range. | Non-recoverable | 19 |
| #3 | A redundant switch was missing its ground sense, or the contacts did not report as expected at startup. | N/A | 15 |
| #23017 | A motor did not respond as expected, and the measured motion did not match the internal stimulation of the motor. | Recoverable | 14 |
| #2 | A reference voltage was out of range. | N/A | 14 |
| #31030 | One of the camera controller units in the doco has failed to power on after multiple attempts or has shut down after initially powering up. | N/A | 14 |
| #5 | One or more fans are not moving as desired | N/A | 10 |
| #297 | An electronic component was reporting an incorrect configuration. | Non-recoverable | 9 |
| #252 | Master supervisory controller did not receive an expected message within a specified time. | Recoverable | 8 |
| #23020 | One of the switches in a specific manipulator is showing inconsistent signals on its two switch leads. | Recoverable | 7 |
| #25589 | During the power up self-test, the remote arm controller board (rac) brakes failed the brake voltage test. | Recoverable | 6 |
| #25588 | A sympathetic error and occurs during the self-test upon system power-up when a loop response test fails. | Recoverable | 6 |
| #23007 | On startup, one or more robotic joints on the manipulator did not make the prescribed test motion to within the specified tolerance. | Recoverable | 6 |
| #21003 | The arm did not perform the commanded motions during startup within a specified tolerance. | N/A | 5 |
| #281 | A processor did not complete a step during system startup within the allotted time. | Non-recoverable | 5 |
| #23034 | After a specified amount of time, a valid event was not seen for one of the remote compute engine switches. | Recoverable | 5 |
| #45049 | A communication timeout with the software running the da Vinci onsite application. | Recoverable | 4 |

## Table 4. Summary of death and injury reports (2000–2012)

| | | |
|---|---|---|
| | **Death Reports (Total = 86)** | |
| | **Example Causes** | **Number of Reports (%)** |
| | Surgeon/staff mistake | 6 (7.0%) |
| | Patient's history | 10 (11.6%) |
| | Inherent risks | 19 (22.1%) |
| | N/A | 27 (31.4%) |
| **During Procedure** | Punctures, bleeding, pulmonary embolism, cardiac arrest | 64 (75.3%) |
| **After Procedure** | Infection/sepsis, heavy bleeding | 15 (17.4%) |
| **Injury Reports (Total = 410)** | | |
| **Example Causes** | | **Number of Reports (%)** |
| Device malfunctions[a] | | 254 (62.0%) |
| Surgeon/staff mistake | | 29 (7.1%) |
| Improper positioning of the patient led to post-operation complications such as nerve damage | | 17 (4.1%) |
| Inherent risks of surgery and patient history | | 16 (3.9%) |
| Burning of tissues near port incisions | | 9 (2.2%) |
| Possible passing of the electrosurgical unit currents through instruments to the patient body | | 6 (1.5%) |
| Surgeon felt shocking at the surgeon-side console | | 2 (0.5%) |
| N/A | | 77 (18.8%) |

[a] Table 5 lists some example reports on device malfunctions that impacted patients during cardiothoracic procedures.

**Table 5. Example malfunctions and their patient impact during cardiothoracic procedures**

| MAUDE Report No. (Year) | Surgery Type | Event | Faulty Component | Patient Impact | Recovery Actions |
|---|---|---|---|---|---|
| 932174 (2007) | Mitral valve repair | Possibly port placement; The robotic arms were never seen to collide, but this could have occurred, resulting in pressure on the retractor. | N/A | Left atrial disrupted, A 3 cm tear occurred in the hood of atrium medial to left atrial appendage, extending down towards the mitral annulus. | A patch was brought into place, trimmed, sewn using a suture, and tied. Sternotomy incision made for further repair. |
| 1077464 (2008) | Beating heart double vessel CABG | Unexplained movement of the system arm with the endowrist stabilizer instrument attached to it. Feet at the distal end of the endowrist stabilizer tipped downward. | System arm | Damage to the myocardium of the patient's left ventricle. | Converted to open sternotomy and repaired. |
| 1590517 (2010) | CABG | Micro bipolar forceps (mbf) instrument jumped forward. When master tool manipulator was moved, the instrument felt stuck and then moved. | Patient side manipulator (psm) | Patient's artery punctured. | Damaged section of artery was transected and the healthy portion was used to complete the bypass. Company replaced the psm component. |
| 2494890 (2012) | CABG | Arcing from bipolar forceps instrument. | Forceps instrument | Small burn to diaphragm. | Connected ground pads and checked electrical surgical unit. |
| 2816230 (2012) | N/A | Patient-side manipulator (psm) arm 2 jumped. | Patient-side manipulator (psm) | The forceps instrument on the psm lacerated the patient's mammary artery. | Converted to open surgery. |

## Table 6. Example complex robotic interactions with possible failure modes

1) A surgeon or surgical assistant needs to be by the patient's side, inserting the ports/scope/instruments.
2) The main surgeon sits at a console some distance away from the patient, with no peripheral vision, and so does not get to see the manipulation of the arms in and around the patient.
3) Any change of instrumentation requires a pause in proceedings, as the patient-side surgeon stops and changes instruments. Once the instrument is docked in the port, registered, and secured, the procedure can be resumed. from where it was stopped. Each of those instrument changes takes about 30 seconds to 2 minutes, so if there are 10 instrument changes in a case, that add 20 minutes to the total time of the procedure.
4) There is no tactile feedback or haptics. Several of the adverse events included inadvertent injury to the aorta, right ventricle, lungs, etc. Sometimes, vessels have been ripped because of lack of feel, and the force delivered by the grasping forceps might significantly exceed safe limits.
5) The endoscope's field of vision is very limited and it can be easy to get disoriented, in terms of both the horizon and the location within the body.
6) Visualization requires insufflation of carbon dioxide at a high flow of 6–10 liters/minute. While $CO_2$ insufflation is also done in non-robotic laparoscopy, it is usually not at such a high flow. That high flow of $CO_2$ can result in absorption of carbon dioxide, which can cause significant metabolic derangements that affect the heart.
7) Each instrument may be used only 10 times, after which software shutdown occurs, driving up the costs and making instruments part of the disposable costs. In open and non-robotic laparoscopic surgeries, some disposable instruments are used, but they are not as expensive as robotic instruments.
8) There is an obligatory setup time, in addition to longer operative times with the robot. Robotic procedures in all fields of surgery take longer than non-robotic (open or laparoscopic) procedures.